\documentclass[runningheads]{llncs}
\UseRawInputEncoding
\usepackage[T1]{fontenc}
\usepackage{graphicx}
\usepackage{algorithm}
\usepackage{algorithmic}
\usepackage{multicol}
\usepackage{multirow}
\usepackage{times}
\usepackage{latexsym}
\usepackage{placeins} 
\usepackage{afterpage} 
\usepackage{listings}
\usepackage{colortbl}
\usepackage{tabularx}
\usepackage{stfloats}
\usepackage{microtype}
\usepackage{tcolorbox}
\usepackage{inconsolata}

\usepackage{hyperref}
\usepackage{float}
\usepackage{booktabs}

\usepackage{amsmath} 
\usepackage{xcolor}  
\usepackage{subdepth} 

\newcommand{\downred}[1]{\textsubscript{\textcolor{red}{$\uparrow$#1}}}
\newcommand{\downgreeen}[1]{\textsubscript{\textcolor{teal}{$\downarrow$#1}}}

\begin{document}
\title{InsQABench: Benchmarking Chinese Insurance Domain Question Answering with Large Language Models}
\titlerunning{InsQABench}
%
\author{Jing Ding\inst{1} \and
Kai Feng\inst{1} \and
Binbin Lin\inst{1}\and
Jiarui Cai\inst{1} \and 
Qiushi Wang\inst{2} \and 
Yu Xie\inst{3}\\
Xiaojin Zhang\inst{1} \and
Zhongyu Wei\inst{2} \and
Wei Chen\inst{1}\thanks{Corresponding author.}}

%
%
\institute{Huazhong University of Science and Technology, Wuhan, China \and
Fudan University, Shanghai, China \and
Purple Mountain Laboratories, Nanjing, China}
\maketitle       
\begin{abstract}

The application of large language models (LLMs) has achieved remarkable success in various fields, but their effectiveness in specialized domains like the Chinese insurance industry remains underexplored. The complexity of insurance knowledge, encompassing specialized terminology and diverse data types, poses significant challenges for both models and users. To address this, we introduce InsQABench, a benchmark dataset for the Chinese insurance sector, structured into three categories: Insurance Commonsense Knowledge, Insurance Structured Database, and Insurance Unstructured Documents, reflecting real-world insurance question-answering tasks.We also propose two methods, SQL-ReAct and RAG-ReAct, to tackle challenges in structured and unstructured data tasks. Evaluations show that while LLMs struggle with domain-specific terminology and nuanced clause texts, fine-tuning on InsQABench significantly improves performance. Our benchmark establishes a solid foundation for advancing LLM applications in the insurance domain, with data and code available at \href{https://github.com/HaileyFamo/InsQABench.git}{InsQABench}.

\end{abstract}

\section{Introduction}

\begin{figure*}
    \includegraphics[width=\linewidth]{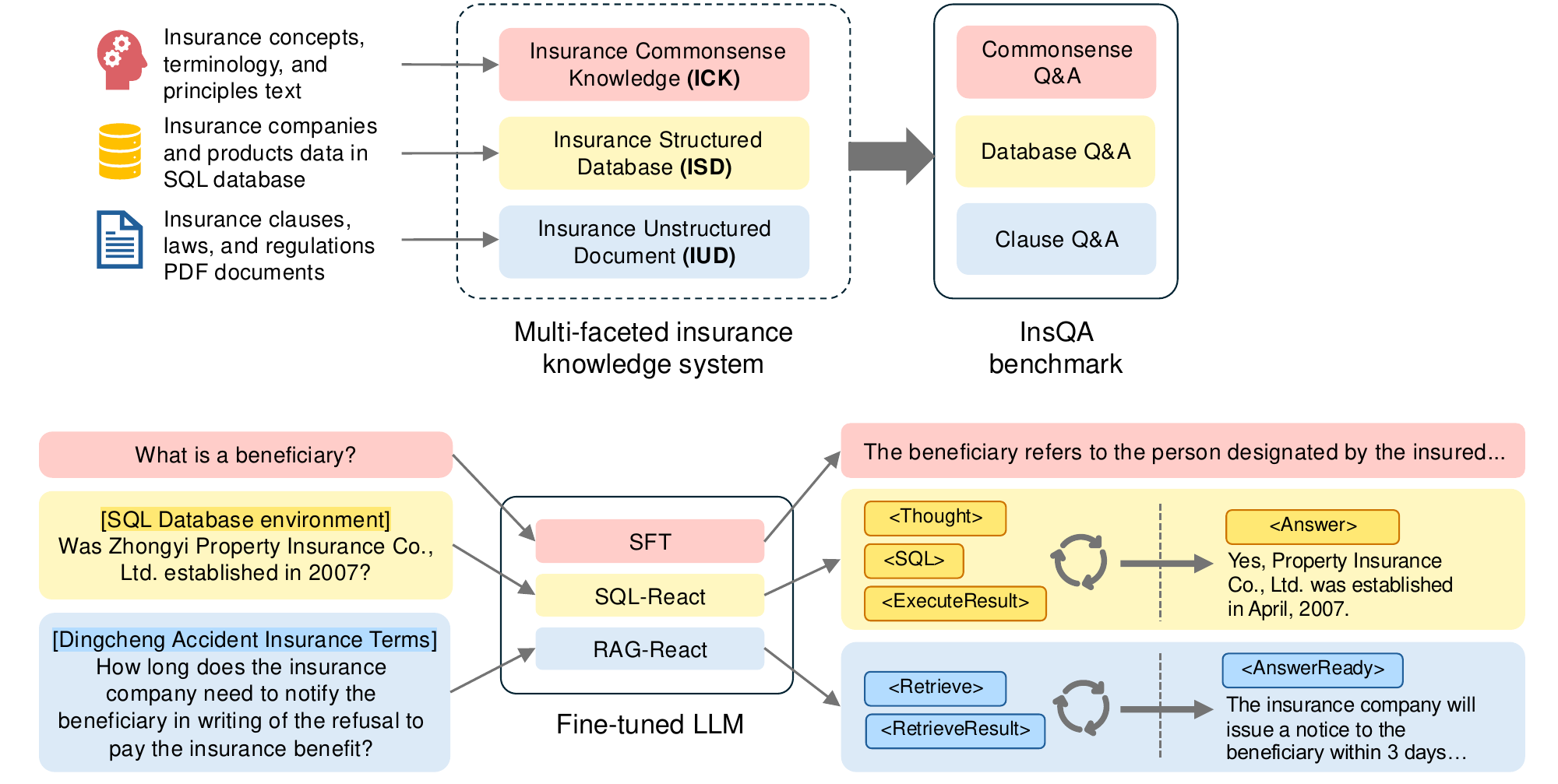}
    \caption{Overview of the InsQABench benchmark, illustrating the multi-faceted insurance knowledge system and fine-tuned LLMs utilizing SQL-ReAct and RAG-ReAct for task-specific enhancements.}
\end{figure*}

The insurance industry plays a critical role in mitigating financial risks and uncertainties, providing essential protection to individuals and businesses alike \cite{liedtke2007s}. However, the complexity of insurance products and the specialized knowledge required to understand them often creates a barrier for the general public. This knowledge gap hinders people from making informed decisions and can contribute to confusion and mistrust towards insurance services \cite{levitt2023complexity,cheston2014deterrents}.

Recent advancements in artificial intelligence, particularly in natural language processing (NLP), have opened new avenues for addressing this challenge. Large language models (LLMs), such as GPT-4~\cite{achiam2023gpt} and Claude-3~\cite{anthropic2024claude3}, demonstrate the capacity to process large volumes of text and generate human-like responses \cite{brown2020language,zhao2023survey}. While promising, applying these models to specialized domains like insurance is far from straightforward. The fragmented nature of insurance data, combined with complex terminologies and ever-changing regulations, presents significant obstacles.

To address the above challenges, we collect a vast amount of insurance-related information from various insurance company official websites and public online resources by a semi-automated approach. We then employ to classify the collected data into three main categories: 1) \emph{Insurance Commonsense Knowledge} (ICK), encompasses essential concepts, terminology, and principles, providing accessible and introductory knowledge for insurance industry newcomers and the general public; 2) \emph{Insurance Structured Database} (ISD), consists of structured information about insurance companies and products, stored in a SQL database, enabling individuals to query and analyze insurance statistics; 3) \emph{Insurance Unstructured Document} (IUD), comprises comprehensive, original information stored in unstructured documents, most in PDF or Word format, such as Insurance Clauses, laws, and regulations, tailored for insurance marketing staffs, product developers, and regulators who require in-depth and authoritative information. ICK represents static knowledge, while ISD and IUD that enables efficient data manipulation are dynamic knowledge. This multi-faceted insurance knowledge system provides a scalable and comprehensive approach to insurance knowledge management.

We have further compiled a comprehensive benchmark dataset, \emph{InsQABench}, which encompasses three specialized insurance question-answering (QA) tasks, each designed to reflect the three types of insurance knowledge:

\emph{Insurance Commonsense QA}, involves foundational insurance concepts, where we leverage supervised fine-tuning (SFT) to improve the model's ability to understand and answer basic insurance-related queries.
\emph{Insurance Database QA}, where the SQL-ReAct method is introduced, enabling models to generate accurate SQL queries, interact with structured insurance databases, and iteratively refine query results for precise answers.
\emph{Insurance Clause QA}, for handling complex legal documents, the RAG-ReAct method is introduced, which allows models to retrieve, interpret, and extract critical information from unstructured data such as insurance clauses, regulations, and legal texts.

Together, these tasks form the foundation of the \emph{InsQABench} benchmark, providing a comprehensive resource for testing and improving the performance of NLP models in the insurance domain.

To address these tasks effectively, we carried out a two-stage evaluation. First, we assessed the impact of supervised fine-tuning (SFT) on the models by comparing their performance on \emph{InsQABench} before and after fine-tuning. This comparison established the foundational improvement gained by adapting models specifically to insurance-related data.

Next, we proposed two task-specific methods tailored to the unique characteristics of our QA tasks:

    \emph{SQL-ReAct}, which equips models with the capability to construct accurate SQL queries, dynamically interact with structured insurance databases, and iteratively refine results for precise answers in database-focused tasks.
    \emph{RAG-ReAct}, tailored for unstructured document QA, which enhances retrieval-augmented generation (RAG) with reactive feedback, enabling models to efficiently retrieve, interpret, and extract critical information from complex legal documents, such as insurance clauses and regulatory texts.

Using these methods, we further evaluated the fine-tuned models on \emph{InsQABench}, comparing their performance against standard approaches like RAG. The results demonstrate that our methods not only yield significant performance gains over baseline methods but also surpass some state-of-the-art proprietary models in the insurance domain.

In summary, our work provides a pioneering benchmark dataset, \emph{InsQABench}, alongside advanced task-specific methodologies, offering a framework for advancing NLP research in the insurance field. By demonstrating the effectiveness of supervised fine-tuning and the proposed methods, this work lays a solid foundation for enhancing the adaptability and reliability of AI systems in specialized, high-stakes domains like insurance.

\section{Related Work}

The emergence of Transformer-based models has catalyzed the development of large language models (LLMs) across various domains, including law \cite{cui2023chatlaw,yue2023disclawllm,xiao2021lawformer,blair2023can}, bio-medicine \cite{singhal2023large,peng2023study,gu2021domain,bao2023disc}, and finance \cite{chen2023disc,li2023large,huang2023finbert}, where they are fine-tuned on domain-specific datasets to achieve notable success. However, in the insurance sector, the adoption of LLMs is still in its early stages.

Existing studies, such as ActuaryGPT \cite{balona2023actuarygpt}, explore the use of LLMs to assist actuaries with tasks like actuarial calculations and report generation. Similarly, many insurance companies have started developing proprietary LLMs, emphasizing internal business needs such as automated sales, customer service, and management. These models integrate proprietary data and industry knowledge but remain tailored to organizational workflows rather than addressing consumer-facing challenges.

In contrast, our work emphasizes user-centric scenarios. We construct a dataset designed to support diverse question-answering tasks related to commonsense insurance knowledge, structured industry databases, and unstructured documents like clauses. By addressing these tasks with methods tailored to the unique characteristics of insurance QA, our approach bridges the gap between general-purpose LLMs and the specialized demands of consumer-facing applications in the insurance domain.

\section{Method}
\label{sec: method}

We tackle three insurance domain QA tasks - Insurance Commonsense QA, Insurance Database QA, and Insurance Clause QA - by first fine-tuning large language models (LLMs) on our InsQABench Dataset using supervised fine-tuning (SFT) with LoRA\cite{hu2021loralowrankadaptationlarge}. Building on this, we propose two tailored methods: SQL-ReAct for structured data tasks and RAG-ReAct for unstructured document tasks.

\subsection{Supervised Fine-Tuning (SFT)}
Using LoRA\cite{hu2021loralowrankadaptationlarge}, we fine-tune pre-trained LLMs on the InsQABench Dataset across all three QA tasks. This step enhances the model's ability to handle insurance-specific terminology and task requirements. Performance comparisons with non-fine-tuned models show that SFT leads to notable improvements in domain-specific QA accuracy, providing a strong baseline for further enhancements. For Database QA and Clause QA, we make some improvements to the fine-tuning method of the model to improve the final performance. The details are shown in \S~\ref{Fine-tune}.

\subsection{SQL-ReAct}

The Insurance database QA task aims to generate a response $r$ given a user's question $q$, the database schema $\mathcal{S}$ and the corresponding database execution engine $\mathcal{E}$. Unlike the typical Text-to-SQL~\cite{katsogiannis2023survey,mohammadjafari2024natural,zhu2024large,liu2024survey,cao2024rsl} task, which focuses on generating a single SQL statement, our task's goal is to generate the final answer. The model must first produce all required SQL statements based on the user's query, then provide the final answer by interpreting the feedback from executing all the SQL statements on the database. The task presents several challenges, including understanding complex query semantics, handling cell mismatches, dealing with SQL execution errors, extracting relevant information from query results, etc. For example, with cell mismatches, the insurance company or product name mentioned in the user's query may not match the cell value in the database (like frequently used abbreviated names), leading to SQL execution errors or null values, thus affecting user experience.

To address above challenges, we propose a baseline method called SQL-ReAct. Inspired by ReAct~\cite{yao2022react}, SQL-ReAct employs an iterative approach to continuously refine SQL statements based on feedback from the database until the desired query result is ready. SQL-ReAct introduces a structured inference process to achieve accurate and relevant responses. This process utilizes specially defined tokens and flags to guide the iterative refinement. Specifically, we adopt a multi-turn dialogue approach to simulate this iterative process.

\textbf{Thought and SQL Generation} \qquad In each iterative step, two key text blocks are generated, i.e., \texttt{<Thought>...</Thought>}, which represents the current generated thought, and \texttt{<SQL>...</SQL>}, which contains the current generated SQL statement. \textbf{The \emph{Thought} block} encapsulates the current reasoning and analysis based on the historical chain-of-thought process. The thought process can encompass various scenarios, such as handling mismatches between entities mentioned in the user's question and the actual values stored in the database tables, identifying and correcting syntax errors in the generated SQL statements, dealing with empty result sets by modifying the query conditions or logic, etc.
 \textbf{The \emph{SQL} block}, on the other hand, wraps the SQL statement generated based on the current thought, which can be either the target SQL statement directly solving the user's question or some auxiliary SQL statements used for observation and assistance. 

\textbf{SQL Execution and Feedback Integration} \qquad Each time, the model can generate the thought process and the statement wrapped in \texttt{<SQL></SQL>} based on the returned results from the database. The SQL statement within the \emph{SQL} block is then extracted and executed by the database engine. The execution results undergo post-processing to avoid appending an excessive amount of data. Specifically, two key pieces of information are extracted from the results: 1) the number of rows returned by the query, and 2) the first $k$ rows of the result set, where $k$ is a predefined constant. These processed results are then appended to all previously decoded content, providing the necessary feedback for further refinement.

\textbf{Iterative Refinement and Termination} \qquad
When the model believes that the previous dialogue results are sufficient to provide an answer to the user, it will respond with an answer wrapped in \texttt{<Answer></Answer>}, at which point the iterative process can be stopped. If the number of iterations exceeds the predefined maximum, an answer will also be forcibly generated.

In \S~\ref{sec:database qa}, we further introduce how to generate training data in the SQL-ReAct approach.

\subsection{RAG-ReAct}
The Insurance Clause QA task focuses on generating accurate responses to user queries about insurance documents. Unlike multimodal document QA datasets such as DocVQA~\cite{mathew2021docvqa}, MP-DocVQA~\cite{mpdocvqa} and DUDE~\cite{dude2023icdar} that contain rich visual elements, insurance clause documents are predominantly text-based. This task presents unique challenges: complex and diverse layout structures, lengthy documents spanning dozens of pages, and abundant domain-specific terminology that creates comprehension barriers for non-expert readers. To address these challenges, we propose RAG-ReAct, a framework that combines rule-enhanced PDF parsing with iterative reasoning to accurately extract and synthesize information from insurance documents while maintaining crucial semantic relationships.

\textbf{PDF Parsing and Dense Retriever} \qquad The process begins with an insurance document, a user question, and a set of tools: a PDF parser, a dense retriever, and an LLM. First, the PDF document is parsed into structured text chunks using the Adobe PDF Extract API~\cite{adobeExtractText}, with custom parsing rules applied to accurately capture complex clauses. The details of the PDF parsing process are further discussed in \S~\ref{sec:Clause data}. These parsed chunks are pre-encoded with a dual-encoder dense retriever model (BGE-M3)~\cite{chen2024bge} and stored in a Faiss vector database, allowing fast and effective retrieval based on semantic similarity with any user query. 

\textbf{Iterative Retrieval Query Generation} \qquad The retrieval process proceeds iteratively. Initially, the user query is sent to the dense retriever to find relevant document chunks. The LLM then uses these chunks to refine the query and retrieve additional chunks in subsequent iterations. If the LLM recognizes an \texttt{<AnswerReady>} signal within its outputs, it indicates that sufficient information has been collected. At this point, the LLM uses the accumulated information to generate a complete answer.

\textbf{Answer and Evidence Generation} \qquad The final answer includes supporting evidence in the form of identifiers for the retrieved text chunks, allowing users to verify the origin of each piece of information. This traceable approach enhances transparency and increases trust in the system's responses.

\section{InsQABench Dataset}
\label{sec:data_construction} 

In this section, we describe the basic information and the construction process of InsQABench Dataset. More details are shown in \S~\ref{appendix:data}.

\begin{table*}[htbp]
\centering
\caption{Overview of the InsQABench Dataset, presenting detailed statistics for each sub-dataset (Commonsense QA, Database QA, and Clause QA) across various metrics, as well as additional source-specific characteristics. The Database QA subset provides structured insurance data, including a number of entries in our custom-built database, with detailed information on tables, columns, data records, keys, and foreign keys. The Clause QA subset covers extensive clause details with average page counts and character counts, highlighting the dataset's richness and comprehensiveness across diverse insurance-related topics.}

\belowrulesep=0pt
\aboverulesep=0pt
\centering
\resizebox{0.7\textwidth}{!}{%
\begin{tabular}{@{}c|c|c|cccc|c@{}}
\toprule
\multirow{6}{*}{\textbf{Commonsense QA}} &
  \multirow{2}{*}{\textbf{Dataset}} &
  \multirow{2}{*}{\textbf{Size}} &
  \multicolumn{4}{c|}{\textbf{Length Distribution}} &
  \multirow{2}{*}{\textbf{Source}} \\ \cmidrule(lr){4-7}
 &                        &                      & \textbf{Type} & \textbf{Mean} & \textbf{Medium} & \textbf{P95} &                                 \\ \cmidrule(l){2-8} 
 & \multirow{2}{*}{Train} & \multirow{2}{*}{10k}  & Question      & 7.43          & 7.00            & 13.00        & \multirow{2}{*}{BX\_GPT\_3.5}     \\
 &                        &                      & Answer        & 206.04        & 198.00          & 328.00       &                                 \\ \cmidrule(l){2-8} 
 & \multirow{2}{*}{Test}  & \multirow{2}{*}{990} & Question      & 7.40          & 7.00            & 13.00        & \multirow{2}{*}{InsuranceQA\_zh} \\
 &                        &                      & Answer        & 202.89        & 194.00          & 326.00       &                                 \\ \bottomrule
\end{tabular}%
}

\resizebox{\textwidth}{!}{%
\begin{tabular}{c|c|l|cccc|c|ccccc}
\toprule
\multirow{6}{*}{\textbf{Database QA}} &
  \multirow{2}{*}{\textbf{Dataset}} &
  \multirow{2}{*}{\textbf{Size}} &
  \multicolumn{4}{c|}{\textbf{Length Distribution}} &
  \multirow{2}{*}{\textbf{Source}} &
  \multicolumn{5}{c}{\textbf{Database Details}} \\ \cmidrule(lr){4-7} \cmidrule(l){9-13} 
 &
   &
   &
  \textbf{Type} &
  \textbf{Mean} &
  \textbf{Medium} &
  \textbf{P95} &
   &
  \textbf{Tables} &
  \textbf{Columns} &
  \textbf{Data Records} &
  \textbf{Keys} &
  \textbf{Foreign Keys} \\ \cmidrule(l){2-13} 
 &
  \multirow{2}{*}{Train} &
  \multirow{2}{*}{44k} &
  Question &14.95
   &14.00
   &26.00
   &
  \multirow{4}{*}{Insurance Clauses} &
  \multirow{4}{*}{2} &
  \multirow{4}{*}{35} &
  \multirow{4}{*}{25620} &
  \multirow{4}{*}{2} &
  \multirow{4}{*}{1} \\
 &                       &                      & Answer   &42.89  & 33.00 &115.00  &  &  &  &  &  &  \\ \cmidrule(lr){2-7}
 & \multirow{2}{*}{Test} & \multirow{2}{*}{546} & Question & 12.89 & 12.00 &22.00  &  &  &  &  &  &  \\
 &                       &                      & Answer   &61.26  & 45.00 & 175.00 &  &  &  &  &  &  \\ \bottomrule
\end{tabular}%
}

\resizebox{\textwidth}{!}{%
\begin{tabular}{@{}c|c|l|cccc|c|cccl@{}}
\toprule
\multirow{6}{*}{\textbf{Clause QA}} &
  \multirow{2}{*}{\textbf{Dataset}} &
  \multirow{2}{*}{\textbf{Size}} &
  \multicolumn{4}{c|}{\textbf{Length Distribution}} &
  \multirow{2}{*}{\textbf{Source}} &
  \multicolumn{4}{c}{\textbf{Source Details}} \\ \cmidrule(lr){4-7} \cmidrule(l){9-12} 
 &
   &
   &
  \textbf{Type} &
  \textbf{Mean} &
  \textbf{Medium} &
  \textbf{P95} &
   &
  \textbf{Avg Pages} &
  \textbf{Max Pages} &
  \textbf{Avg Chars} &
  \textbf{Max Chars} \\ \cmidrule(l){2-12} 
 &
  \multirow{2}{*}{Train} &
  \multirow{2}{*}{40k} &
  Question &
  15.10 &
  14.00 &
  26.00 &
  \multirow{4}{*}{Insurance Clauses} &
  \multirow{4}{*}{5.19} &
  \multirow{4}{*}{66.00} &
  \multirow{4}{*}{4815.44} &
  \multirow{4}{*}{53921.00} \\
 &                       &                      & Answer   & 288.67 & 265.00 & 527.00 &  &  &  &  &  \\ \cmidrule(lr){2-7}
 & \multirow{2}{*}{Test} & \multirow{2}{*}{870} & Question & 13.05  & 12.00  & 22.00  &  &  &  &  &  \\
 &                       &                      & Answer   & 251.56 & 227.00 & 456.19 &  &  &  &  &  \\ \bottomrule
\end{tabular}%
}
\end{table*}

\begin{figure}
    \centering
    \includegraphics[width=\linewidth]{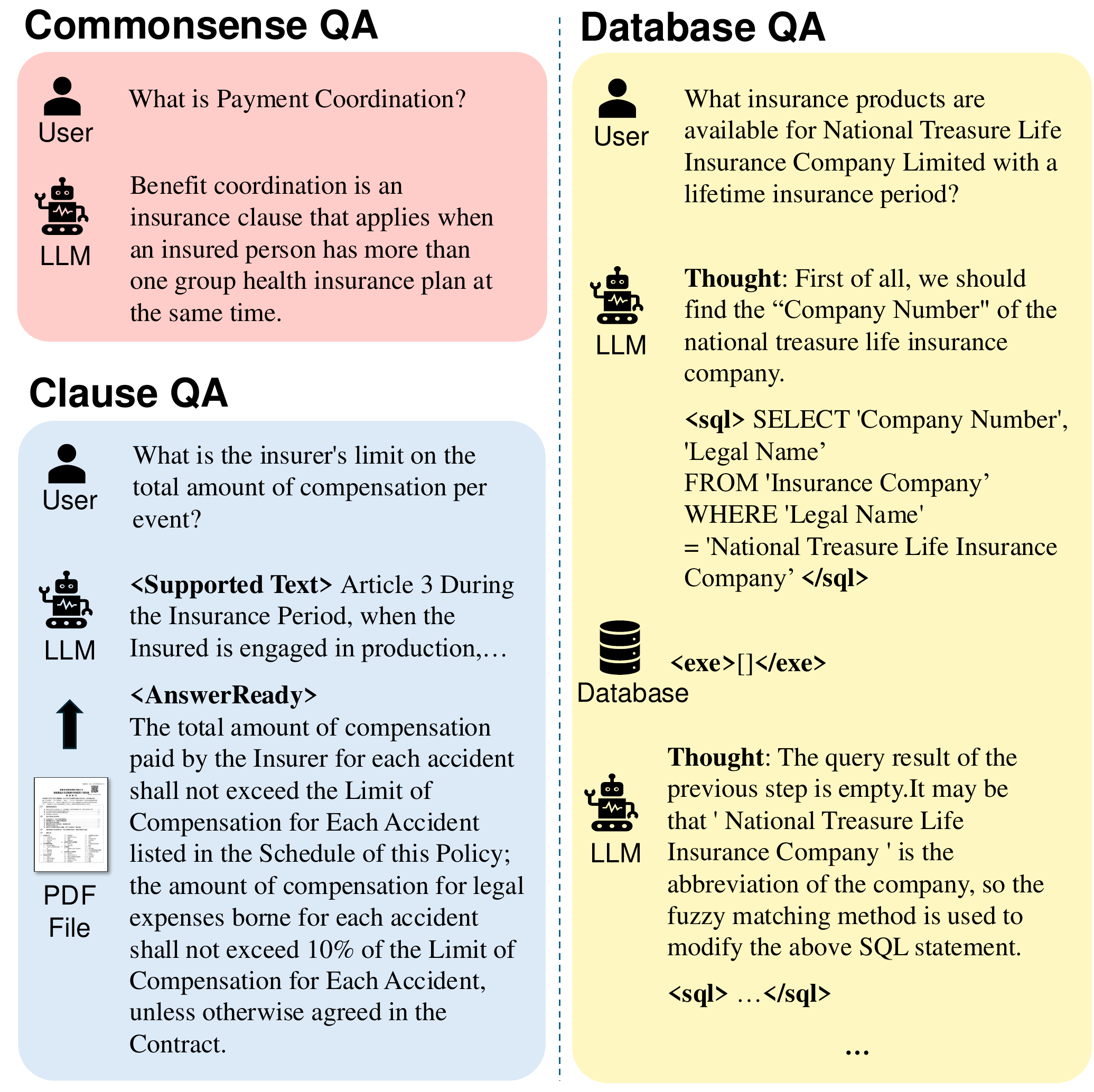}
    \caption{Examples in the InsQABench Dataset.}
    \label{fig:enter-label}
\end{figure}

\subsection{Insurance Commonsense QA}

To construct high-quality insurance domain commonsense question-answering dataset that aligns with real-world applications, we collect training and test sets from various sources.

\paragraph{Training Set for Commonsense QA} We first crawl 8k insurance-related questions posted by users on a popular online QA forum\footnote{\url{https://zhidao.baidu.com/}}, along with their corresponding highest-voted answers provided by the community. To ensure the quality of answers, we utilize GPT-3.5\cite{openai2023chatgpt} to refine the original responses. The raw question-answer pairs are used as input, and the model generates optimized answers that improve the professionalism, readability, and comprehensibility. In addition, we engage professional insurance domain experts to manually compose approximately 2k additional question-answer pairs. The experts are instructed to focus on crafting high-quality, canonical answers to common insurance questions frequently posed by novice users. The resulting dataset serves as the training set for this task, yields a total of 10k samples.

\paragraph{Test Set for Commonsense QA} For the test set, we adopt the test set of InsuranceQA dataset\footnote{\url{https://github.com/shuzi/insuranceQA}}, which is, to the best of our knowledge, the only publicly available Chinese insurance question-answering dataset. We sample 990 QA pairs, with questions from genuine users and high-quality answers manually written by insurance domain experts, ensuring the dataset's high representativeness of real-world scenarios and its value for evaluating the performance of insurance-domain question-answering systems.

\subsection{Insurance Database QA}
\label{sec:database qa}
\begin{figure*}[htbp]
    \centering
    \small
    \includegraphics[width=\linewidth]{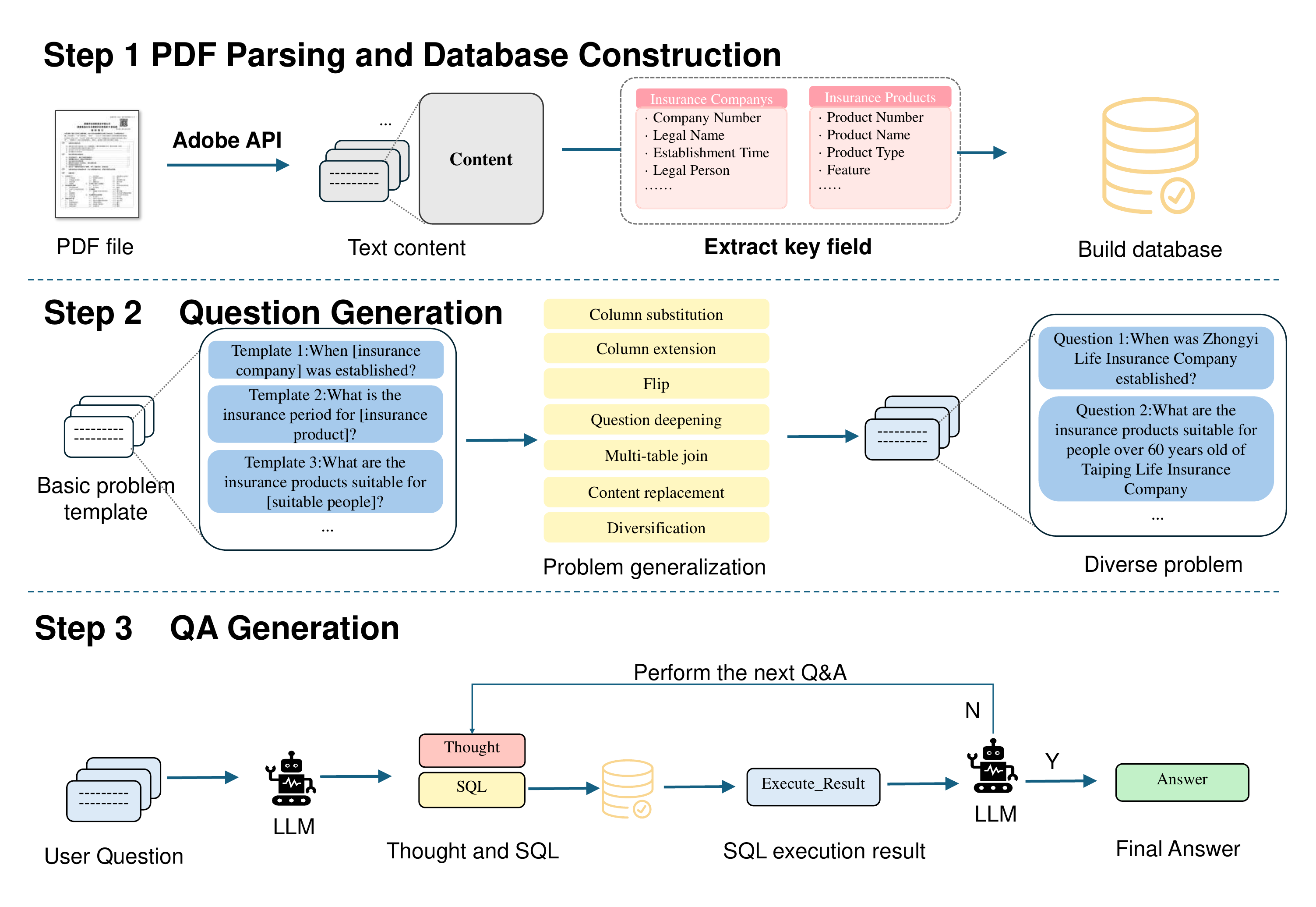}
    \caption{The construction process of the Database QA Dataset.}
\end{figure*}

For the Insurance Database QA task, we first introduce the construction of our insurance database and then describe the method used to build the training and test sets.

\paragraph{Database Construction} Our insurance database is a comprehensive repository of information on insurance companies and their products. It is designed to support users in understanding various aspects of insurance, such as product offerings, features, and statistics. To populate this database, we crawl official websites of 192 insurance companies and collect meta-information and clause documents for 25k insurance products. We employ GPT-3.5 to automatically extract the majority of the information from the website source code and clause documents. To ensure the accuracy and completeness of the database, human annotators thoroughly review the extracted data, identifying and correcting any errors or inconsistencies. For missing fields, the annotators manually fill in the required information. 

\paragraph{Training Set for Database QA} ~ {Inspired by the Evol-Instruct\cite{evol-instruct} method, we propose an approach to enhance the diversity of database query questions. We begin by creating initial question templates using a rule-based method tailored for database question-answering tasks.
To further enrich and diversify these questions, we expand them across seven distinct aspects detailed in Table \ref{tab:question_types}. We then apply an evolutionary process with equal probability for each evolutionary direction, performing five rounds of evolution on the initial templates. If a new question does not introduce additional information, it is not answered but rather moved to the next iteration dataset for further evolution. To ensure practical applicability, we use Gemini 1.5 Pro\cite{team2023gemini} to filter out questions that are too complex to yield results from the database and to rephrase questions into more colloquial language, making them closer to real user language.}

\begin{table}[htbp]
    \centering  
    \caption{The table shows seven ways to generalize the template questions in the Database QA Dataset.} %
\resizebox{\textwidth}{!}{%

    \begin{tabular}{cp{10cm}}   
    \toprule  
    \textbf{Question Type} & \textbf{Description} \\   
    \midrule  
    Column substitution & The original question usually involves a query for a column, and we can replace the column we want to ask for, converting the question to the new column. \\   
    \midrule
    Column extension & Adds columns to the query for the original question. This amounts to an increase in the number of questions for the original question. \\ 
    \midrule
    Flip & Reverse the original question. \\ 
    \midrule
    Content replacement & We randomly extract some data rows from the database, and replace the relevant field content extracted with the corresponding content in the original question. \\ 
    \midrule
    Question deepening & Instead of simply asking for the content of a field, the entity is asked if some condition is met in that field. \\
    \midrule
    Multi-table join & Transform the original problem into a problem of joining multiple tables. \\ 
    \midrule
    Diversification & Taking the original problem and diversifying it into any form. \\ 
    \bottomrule  
    \end{tabular} 
}
    \label{tab:question_types}  
\end{table}

For generating answers, inspired by existing multi-agent based system~\cite{fan2024ai,wei2024mc} and data generation method~\cite{ni2024mechagents}, we employ a dialogue process involving two agents: a virtual user and a database execution tool. When Gemini first receives a query from a user, we provide Gemini with database schema information and guidance information. In each round, Gemini generates a thought process based on the feedback from database, analyzes the problem, and generates SQL statements to solve the problem. The virtual user, represented by the database execution tool, executes these SQL statements and returns the results to Gemini. Gemini evaluates whether the result is sufficient to generate a direct answer or if further iteration is necessary. This dialogue process continues until the SQL statement can be executed successfully to produce the correct answer. Once Gemini determines the database has enough feedback to answer the user's question, it prints Ready. The interactions between the two agents are incorporated into the training dataset, ensuring that the generated answers are accurate and contextually relevant. 

\paragraph{Test Set for Database QA} ~{For the test set, to prevent data contamination contamination, we add new data to the database during testing. The test set is generated only from new data. In order to improve the quality of the test set, we manually compiled 546 test samples covering various types of database questions in the insurance domain and provided standard answers.}

\subsection{Insurance Clause QA}
\begin{figure*}[!h]
    \small
        \includegraphics[width=\linewidth]{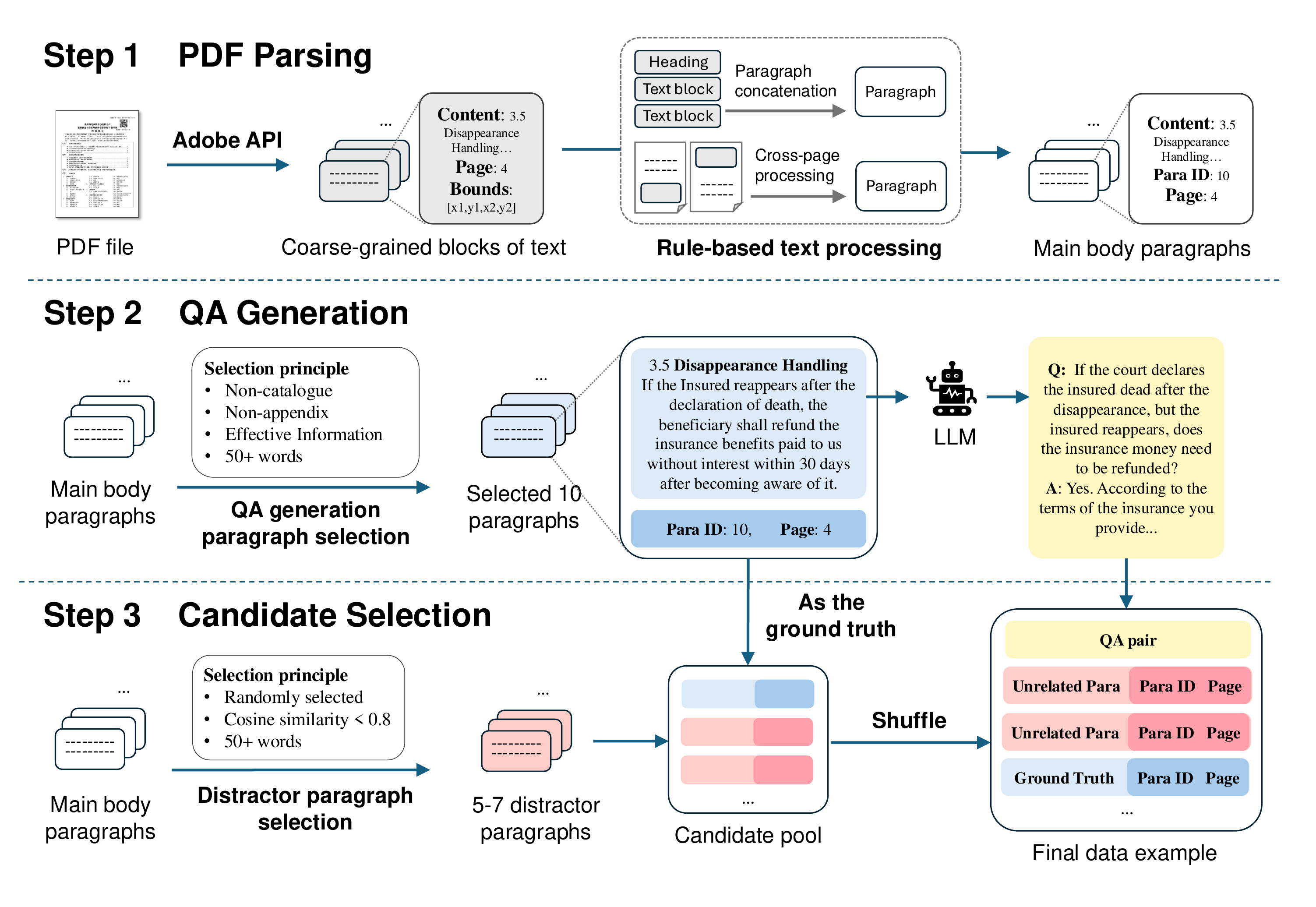}
    \caption{The construction process of the Clause QA Dataset.}
    \label{fig: clause construction}
\end{figure*}

\label{sec:Clause data}
\paragraph{PDF Parsing} ~{We first use the Adobe PDF Extract API to parse the clause PDFs, obtaining element-level text segmentation (as illustrated in Figure \ref{fig: clause construction}). This process retrieves the text and the bounding box coordinates for each element fragment. Following this, we design specific rules tailored to the clause layouts of different companies. These rules leverage the bounding box information to merge the element-level text into coherent paragraph-level segments, enabling us to capture more accurate paragraph text across various clause layouts.}

\paragraph{Training Set for Clause QAE}~{We randomly select 10 paragraphs from a clause. To distinguish it from the Database QA task, we skip the table of contents and company information at the beginning of the clause and focus on the unique textual content of the clause. Also, we omit the appendix at the end of the clause, as most of it is mainly an explanation of expertise in other non-insurance areas. We limit each paragraph to a minimum of 50 Chinese characters to ensure it contains sufficient information.

Next, we feed each selected paragraph into Gemini and prompt it to generate question-answer pairs (QA pairs) for it. At this step, Gemini is prompted to discard paragraphs that are meaningless or have too little information to further ensure the quality of the paragraphs. The number of QA pairs generated depends on the length of the paragraph, varying from 1 to 3. In order to make the generated QA pairs more relevant to the real case of inquiry, we employ a one-shot approach, where Gemini is provided with an example to guide it in generating colloquial questions. The generated questions are prompted to avoid technical nouns to make them more closer to the user's language habits, and better train the model's ability to retrieve the correct answer through fuzzy questions.

To construct a QAE pair, each QA pair is paired with 5-7 candidate paragraphs. These candidates include the selected input paragraph as the ground truth, along with other unrelated paragraphs randomly selected from the whole clause, except for those paragraphs where the cosine similarity to the ground truth is too high. 

The candidates are then shuffled and labeled with `id's, which indicate their initial orders in the clause. We mark the id of the ground truth paragraph in the dataset. 

Finally, we filter out low-quality answers such as questions or incomplete answers, garbled responses, etc. It is important to note that we have retained some of the questions that Gemini reports as "unanswerable", accounting for 5\% of the total data, to simulate questions that cannot be accurately and comprehensively answered by relying on a single clause alone in a real QA scenario. }

\paragraph{Test Set for Clause QAE}~{We first split 870 samples from the Clause QAE Dataset as the initial test set. Then, we manually select 20 PDF documents and carefully review their content. For each document, we choose 5 representative questions, resulting in a high-quality subset of 100 questions. To ensure diversity, the selected questions stem from the clauses of multiple insurance companies and cover a wide range of categories, including interpretation of terms, recommended products, and case studies, etc.}

\section{Experiments}

\subsection{Insurance Commonsense QA task} 

\begin{table}[h]
    \centering
    \begin{minipage}{0.45\textwidth}
        \centering
        \caption{The rule-based evaluation results on Insurance Commonsense QA task.}
        \resizebox{\textwidth}{!}{
        \begin{tabular}{llll} 
            \toprule
            \textbf{Model} & \textbf{Prec.} & \textbf{F1} & \textbf{ROUGE-L}\\
            \midrule
            Baichuan2 \scriptsize{13B} & 38.15 & 36.27 & 7.37 \\
            \rowcolor[HTML]{F2F3F5}
            Baichuan2 (Fine-tuned) \scriptsize{13B} & 46.35\downred{8.2} & 44.42\downred{8.15} & 8.58\downred{1.21} \\
            GLM4 \scriptsize{9B} & 30.57 & 33.88 & 8.77 \\
            \rowcolor[HTML]{F2F3F5}
            GLM4 (Fine-tuned) \scriptsize{9B} & \textbf{47.99}\downred{17.42} & \textbf{45.60}\downred{11.72} & \textbf{9.24}\downred{0.47} \\
            Qwen1.5 \scriptsize{14B} & 32.48 & 34.94  & 8.09 \\ 
            \rowcolor[HTML]{F2F3F5}
            Qwen1.5 (Fine-tuned) \scriptsize{14B} & 47.85\downred{15.37} & 45.19\downred{10.25} & 8.43\downred{0.34} \\
            \bottomrule
        \end{tabular}
        }
        \label{tab:Rule-based Commonsense}
    \end{minipage}
    \hfill
    \begin{minipage}{0.52\textwidth}
        \centering
        \caption{The model-based evaluation results on Insurance Commonsense QA task.}
        \resizebox{\textwidth}{!}{
        \begin{tabular}{lllll} 
            \toprule
            \textbf{Model} & \textbf{ACC} & \textbf{PRO} & \textbf{SIM} & \textbf{AVG}\\
            \midrule
            Baichuan2 \scriptsize{13B} & 64.86 & 64.82 & 67.46 & 65.71\\
            \rowcolor[HTML]{F2F3F5}
            Baichuan2 (Fine-tuned) \scriptsize{13B} & 65.62\downred{0.76} & 65.79
            \downred{0.97} & 68.04\downred{0.58} & 66.48\downred{0.77}\\
            GLM4 \scriptsize{9B} & 64.40 & 66.50 & 68.79  & 67.23\\
            \rowcolor[HTML]{F2F3F5}
            GLM4 (Fine-tuned) \scriptsize{9B} & \textbf{70.26}\downred{5.86} & \textbf{70.41}\downred{3.91} & \textbf{72.57}\downred{3.78} & \textbf{71.08}\downred{3.85} \\
            Qwen1.5 \scriptsize{14B} & 64.12 & 64.26 & 66.42 & 64.93 \\ 
            \rowcolor[HTML]{F2F3F5}
            Qwen1.5 (Fine-tuned) \scriptsize{14B} & 65.90\downred{1.78} & 65.80\downred{1.54} & 68.19\downred{1.77} & 66.63\downred{1.70}\\
            \bottomrule
        \end{tabular}
        }
        \label{tab:Model-based Commonsense}
    \end{minipage}
\end{table}

\paragraph{Evaluation Metric} ~{We initially used a rule-based evaluation system that measures Precision, F1, and ROUGE-L. However, due to the nuanced nature of insurance question-answering, we found that rule-based metrics alone were insufficient for capturing the full depth of model performance. To address this, we introduced a model-based evaluation, scoring the responses across three dimensions: accuracy (ACC), professionalism (PRO), and similarity (SIM) to the reference answers. 
The full score of each dimension is set at 100 points, which are added together to obtain the total score, and the average value is finally taken as the final result.}

\paragraph{Experiment Setup} 
We evaluate Baichuan2-13B-Chat\cite{baichuan2023baichuan2}, GLM4-9B-Chat\cite{glm2024chatglm}, and Qwen1.5-14B-Chat\cite{qwen}, before and after fine-tuning on the Commonsense Dataset using LoRA. Rule-based evaluation is conducted on 990 test samples, from which 100 samples are randomly selected for model-based evaluation scored by GPT-4o\footnote{\url{https://openai.com/index/hello-gpt-4o/}}. The anonymized outputs of these 100 samples are assessed for Accuracy (ACC), Professionalism (PRO), and Similarity (SIM).

\paragraph{Results and Analysis}
In the rule-based evaluation (Table \ref{tab:Rule-based Commonsense}), fine-tuned models consistently outperform their base counterparts in all metrics. GLM4-9B-Chat achieves the highest performance. The generally low ROUGE-L scores can be attributed to the brevity of responses in the test set, which limits the scope for generating longer, detailed answers. In the model-based evaluation (Table \ref{tab:Model-based Commonsense}), fine-tuned GLM4-9B-Chat again exhibits the most significant improvement in all evaluation dimensions. These results underscore the effectiveness of fine-tuning on our dataset in capturing domain-specific nuances and improving performance across diverse QA tasks.

\subsection{Insurance Database QA task}
\paragraph{Evaluation Metric}
To evaluate the performance of our model on Database QA task, we use multi-dimensional evaluation criteria, combining a subjective scoring method using GPT-4o and automated evaluation metrics. GPT-4o is used to score the accuracy of answers generated by different models, check whether the answers correctly address the user's questions and are consistent with the database content. For automated metrics, we use the ROUGE-1 score to evaluate the lexical overlap between the generated response and the reference answer, and the ROUGE-L score to evaluate the longest common subsequence between the two.

\paragraph{Experiment Setup} 
We test Wenxin\cite{yiyan}, ChatGPT-3.5, Baichuan2-13B-Chat and Qwen1.5-14B-Chat models using our two rounds method and Prompt based SQL-ReAct method, respectively. In addition, we also tested Baichuan2-13b-Chat and Qwen1.5-14B-Chat based on the fine-tuning SQL-ReAct method. In two rounds method, we instruct the model to generate and execute all the SQL statements, and then retrieve the answer. The test data we used came from our test set, which consists of 546 data examples.

\paragraph{Result and Analysis}
Table \ref{tab:iteration} shows the results of the iterative framework and the general QA, respectively. Comparing results in Table \ref{tab:iteration}, we can clearly see the improvement in the performance of our iterative framework for database tasks. After fine-tuning, our model shows significant improvement over the base model and outperforms many closed-source models such as ChatGPT-3.5.

\begin{table}[htbp]
    \centering
    \caption{Evaluation results of the iterative framework on the Database QA task. The two rounds approach allows the model to generate all relevant SQL statements at once and then answer based on the results returned. We then use the SQL-ReAct framework to get evaluation results on these models. Finally, we used our SQL-ReAct framework to fine-tune two base models to get the evaluation results.The performance improvement is compared with the corresponding base model.}

    \begin{minipage}{0.7\textwidth}
    \centering
    \resizebox{\textwidth}{!}{
    \begin{tabular}{llll}
    \toprule
        \textbf{Model} & \textbf{ACC}  & \textbf{ROUGE-1} & \textbf{ROUGE-L} \\
    \midrule
        Wenxin(Two rounds) & 24.35 & 28.49 & 21.08 \\
        \rowcolor[HTML]{F2F3F5}
        Wenxin + SQL-ReAct (w/o. finetuningd)& 23.88\downgreeen{0.47} & 34.35\downred{5.86} & 24.69\downred{3.61} \\
        GPT-3.5(Two rounds) & 33.15 & 32.71 & 25.37 \\
        \rowcolor[HTML]{F2F3F5}
        GPT-3.5 + SQL-ReAct (w/o. finetuning)& 50.73\downred{17.58} & 43.94\downred{11.23} & 34.35\downred{8.98} \\
        Baichuan2(Two rounds) \scriptsize{13B} & 4.89 & 19.63 & 15.13 \\
        \rowcolor[HTML]{F2F3F5}
        Baichuan2 + SQL-ReAct (w/o. finetuning)\scriptsize{13B}  & 12.92\downred{8.03} & 30.91\downred{11.28} & 23.72\downred{8.59} \\
        \rowcolor[HTML]{F2F3F5}
        Baichuan2  + SQL-ReAct (Fine-tuned) \scriptsize{13B} & 52.50\downred{47.61} & 26.52\downred{6.89} & 18.79\downred{3.66} \\
        Qwen1.5(Two rounds) \scriptsize{14B} & 35.27 & 37.06 & 29.03 \\
        \rowcolor[HTML]{F2F3F5}
        Qwen1.5 + SQL-ReAct (w/o. finetuning)\scriptsize{14B}  & 43.23\downred{7.96} & 39.21\downred{2.15} & 30.21\downred{1.18} \\
        \rowcolor[HTML]{F2F3F5}
        Qwen1.5  + SQL-ReAct(Fine-tuned)\scriptsize{14B} & \textbf{57.41} \downred{22.14}& \textbf{49.85}\downred{12.79} & \textbf{39.09}\downred{10.06} \\ 
    \bottomrule
    \end{tabular}
    }
    \end{minipage}
    \hfill

    \label{tab:iteration}
\end{table}

\subsection{Insurance Clause QA task}

\paragraph{Evaluation Metric} We apply similar rule-based and model-based evaluation methods here as in the Commonsense QA task. The rule-based metrics remain consistent. For the model-based evaluation, we adopt three new dimensions: \emph{Accuracy (ACC), Completeness (CPL), and Clarity (CLR)}, inspired by DISC-LawLLM\cite{yue2023disclawllm}. These dimensions aim to measure how well models understand and communicate complex insurance clause information.

\paragraph{Experiment Setup} 
For the rule-based evaluation, we test models on the 870 test examples. For the model-based evaluation, to better approximate real-world usage scenarios, we use the 100 questions from 20 PDFs. Open-source models are provided with the full extracted text from the PDFs, while closed-source models directly receive the original PDF files for processing.
All model outputs are anonymized and submitted together to GPT-4o for evaluation, ensuring unbiased scoring. 

\begin{table}
    \centering
    \begin{minipage}[t]{0.43\textwidth} 
        \caption{The rule-based evaluation results on Insurance Clause QA task.}
        \resizebox{\textwidth}{!}{
        \begin{tabular}{llll}
            \toprule
            \textbf{Model} & \textbf{Prec.} & \textbf{F1} & \textbf{ROUGE-L} \\ 
            \midrule
            Baichuan2 \scriptsize{13B} & 44.30 & 44.06 & 26.98\\
            \rowcolor[HTML]{F2F3F5}
            Baichuan2 (Fine-tuned) \scriptsize{13B} & 59.75\downred{15.45} & 46.54\downred{2.48} & 44.09\downred{17.11}\\
            GLM4 \scriptsize{9B} & 62.01 & 56.42 & 54.34 \\
            \rowcolor[HTML]{F2F3F5}
            GLM4 (Fine-tuned) \scriptsize{9B} & \textbf{76.40}\downred{14.39} & \textbf{69.10}\downred{12.68} & 63.46\downred{9.12} \\
            Qwen1.5 \scriptsize{}{14B} & 58.08 & 45.17 & 31.95 \\
            \rowcolor[HTML]{F2F3F5}
            Qwen1.5 (Fine-tuned) \scriptsize{14B} & 73.28\downred{15.2} & 72.66\downred{27.49} & \textbf{87.20}\downred{55.25}\\
            \bottomrule
        \end{tabular}
        }
        \label{tab:Auto Clause}
    \end{minipage}
    \hfill 
    \begin{minipage}[t]{0.55\textwidth} 
        \caption{The model-based evaluation results on the Insurance Clause QA task.}
        \resizebox{\textwidth}{!}{
        \begin{tabular}{lllll}
            \toprule
            \textbf{Model} & \textbf{ACC} & \textbf{CPL} & \textbf{CLR} & \textbf{AVG} \\
            \midrule
            GPT-4 & 75.41 & 75.77 & 77.51 & 76.23 \\
            Kimi & 78.63 & 79.48 & 79.76 & 79.29 \\
            ChatPDF & 67.29 & 68.46 & 71.06 & 69.60 \\
            Wenxin & 77.79 & 78.64 & 78.80 & 78.41 \\
            GLM4 (Fine-tuned) + RAG \scriptsize{9B}  & 74.10 & 72.82 & 74.00 & 73.64 \\ 
            \rowcolor[HTML]{F2F3F5}
            GLM4 (Fine-tuned) + RAG-ReAct \scriptsize{9B}  & \textbf{83.74}\downred{9.64} & \textbf{83.27}\downred{10.45} & \textbf{83.87}\downred{9.87} & \textbf{83.63}\downred{9.99} \\ 
            Qwen1.5 (Fine-tuned) + RAG \scriptsize{14B} & 72.62 & 72.63 & 73.24 & 72.83 \\ 
            \rowcolor[HTML]{F2F3F5}
            Qwen1.5 (fine-tuned) + RAG-ReAct \scriptsize{14B}  & 73.64\downred{1.02} & 71.73\downgreeen{0.9} & 73.81\downred{0.57} & 73.06\downred{0.23} \\ 
            \bottomrule
        \end{tabular}
        }
        \label{tab:Subjective Insurance Clause}
    \end{minipage}
\end{table}

\paragraph{Results and Analysis}  
The rule-based evaluation results in Table \ref{tab:Auto Clause} highlight the benefits of fine-tuning on the Clause QA dataset. Among the open-source models, the fine-tuned GLM4-9B-Chat achieves the highest F1 (69.10) and strong Precision (76.40), while the fine-tuned Qwen1.5-14B-Chat excels in ROUGE-L (87.20). 
In the model-based evaluation (Table \ref{tab:Subjective Insurance Clause}), GLM4-9B-Chat fine-tuned with RAG-ReAct achieves the best overall performance, outperforming many closed-source models. This demonstrates that our RAG-ReAct method enhances the model's ability to process and understand clause-related queries. 

\section{Conclusion}
We introduce InsQABench, a comprehensive benchmark for Chinese insurance question-answering, covering Commonsense QA, Database QA, and Clause QA. To tackle the complexity of insurance knowledge, we propose SQL-ReAct and RAG-ReAct for structured and unstructured data tasks. Fine-tuning LLMs on InsQABench significantly improved their ability to handle domain-specific terminology and complex clause documents, providing a solid foundation for advancing insurance-specific NLP applications.

\clearpage
\bibliographystyle{splncs04}
\bibliography{refs}

\clearpage
\appendix
\label{appendix}
\section{InsQABench Dataset Details}
\label{appendix:data}
Here we present more details of InsQABench. You can also get our data at \href{https://github.com/HaileyFamo/InsQABench.git}{InsQABench}.

\subsection{Diversity Design}

\label{dataset exp}
We constructed a variety of data for each dataset. In this section, we will show the diversity of data in the sub-datasets.\\

\subsubsection{Database QA Dataset}

\textbf{Schema of the Tables} \\
The schema of the tables in the Database QA Dataset is shown in Tables \ref{tab:company} and \ref{tab:products}. 
Table \ref{tab:company} is the pattern of the insurance company table, and Table \ref{tab:products} is the pattern of the insurance product table.

\begin{table*}[htbp]
    \centering  
    \caption{Table of insurance company in the Database QA Dataset.} %
    \begin{tabular}{cp{10cm}}
    \hline  
    \textbf{Field name} & \textbf{Paraphrase} \\   
    \hline  
    Company number  & Primary key written for each company. \\  
    Legal name & Full legal name of the company. \\    
    Establishment time & When the company was established.\\
    Legal person & Legal representative of the company. \\ 
    Official website & URL of the company's official website. \\
    Corporate domicile & Address of the company's domicile. \\
    Registered capital & The registered capital of the company,
    the unit is 100 million yuan.\\
    Business scope & The scope of the company's insurance business. \\
    Company abbreviation & Short for company.\\
    Type of company & Classification of the company based on its business structure or legal form.\\
    Subsidiary Number & The number of the company to which it is affiliated.\\
    Company switchboard & The main telephone number used for general inquiries at the company’s head office.\\
    Operating area &  The geographic regions or markets where the company conducts its business activities.\\
    Customer service hotline & The phone number designated for customer support and inquiries.\\
    Fax & The number used for sending and receiving faxes.\\
    Postcode & The postal code for the company's address.\\
    Business premises & The location or physical address where the company's business operations are conducted.\\
    \hline  
    \end{tabular}  
    \label{tab:company}  
\end{table*}

\begin{table*}[htbp]
    \centering  
    \caption{Table of insurance products in the Database QA Dataset.} %
    \begin{tabular}{cp{10cm}}   
    \hline  
    \textbf{Field name} & \textbf{Paraphrase} \\   
    \hline  
Product number  & Unique identifier assigned to each insurance product. \\   
    Product name & The official name of the insurance product as listed in company records. \\ 
    Product type & The category or classification of the insurance product.\\
    Feature & Key characteristics or attributes of the insurance product.\\
    Suitable population & The target demographic or group of individuals for whom the insurance product is intended.\\
    Product website & URL of the webpage dedicated to providing information about the insurance product.\\
    Exemption from liability & Conditions under which the insurer is not liable for certain claims or damages.\\
    Deductible amount & The amount that the policyholder must pay out-of-pocket before the insurance coverage kicks in.\\
    Insurance period & The duration for which the insurance policy provides coverage, typically expressed in years.\\
    Waiting period & The time period during which no claims are accepted after the policy begins.\\
    Cooling-off period & The period during which the policyholder can cancel the policy and receive a refund, if applicable.\\
    Insurance liability & The extent of the insurer's financial responsibility in the event of a claim.\\
    Payment method/Insurance age & The available options for premium payment and the age at which the insurance policy is applicable.\\
    Company number & Unique identifier assigned to the insurance company offering the product.\\
    Sales status & The current availability and market status of the insurance product .\\
    Bonus & Additional benefits or rewards offered as part of the insurance policy.\\
    Policy loan & The option for policyholders to take out a loan against the cash value of their insurance policy.\\
  \hline  
    \end{tabular}  
    \label{tab:products}  
\end{table*}

\clearpage
\label{Qustion type}

\begin{figure*}[htbp]
	\centering
    \includegraphics[width=\linewidth]{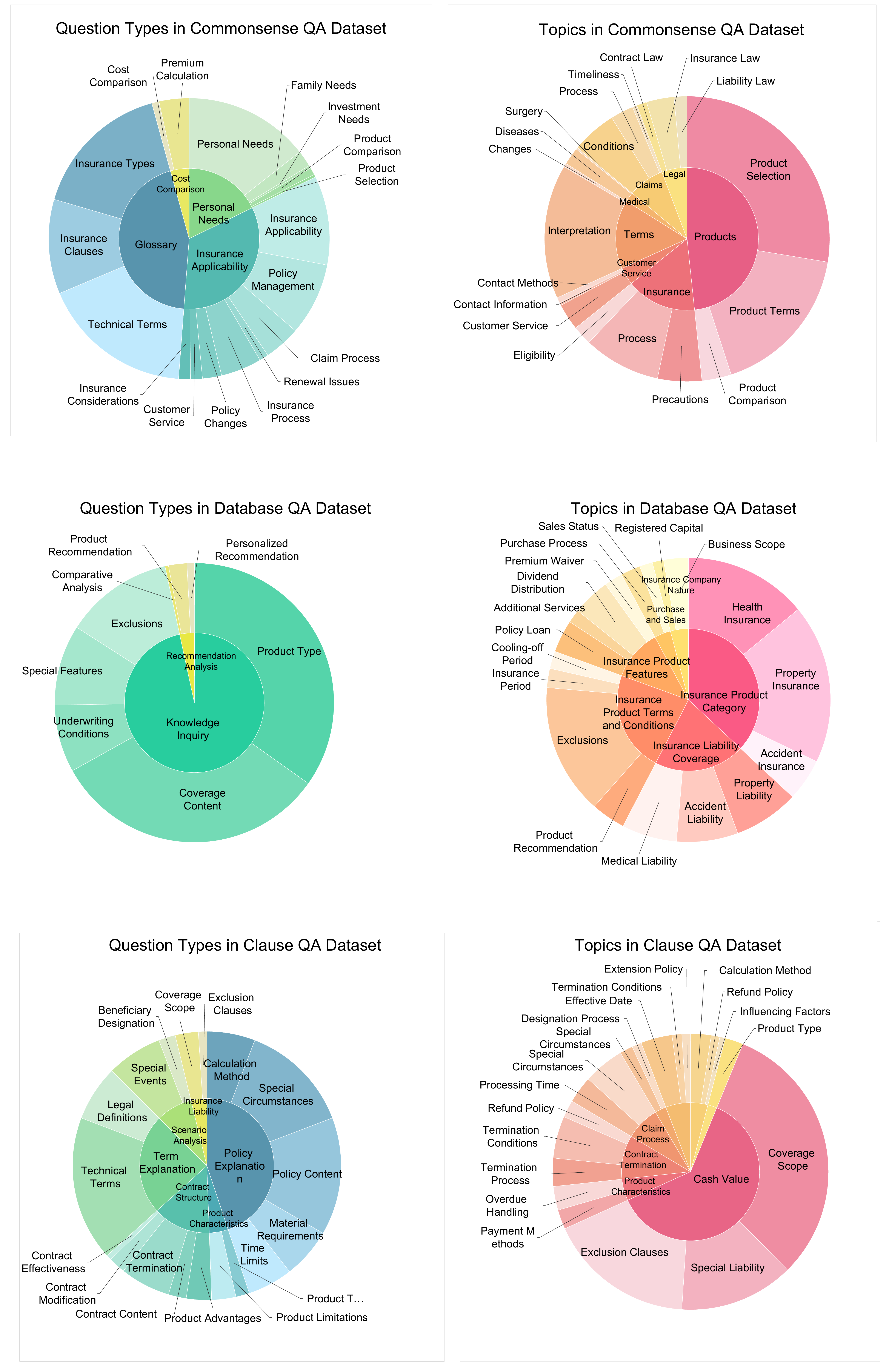}
    \caption{The types and topics of the questions in the three tasks.}
\end{figure*}

\clearpage
\section{Model Fine-tuning}
\subsection{Method}
\label{Fine-tune}
In this part, we present  the improved fine-tuning methods of Database QA and Clause QA based on LoRA.

\paragraph{Database QA} The fine-tuning of the database question answering task is designed to enhance LLMs' ability to find the data from the database that users want to query. For each piece of data in the constructed fine-adjustment dataset, the input sequence of the model is denoted as $x_i$, and the expected output of the model is denoted as $y_i$. 
\begin{equation}
    L(D_{\text{pair}}) = -\log p(y_i | x_i)
\end{equation}
After fine-tuning, the model can output each round of thought markup, thought process, and corresponding SQL statements as expected.

\paragraph{Insurance Clause QA} We use the Clause QAE pairs to train the models. For one QAE pair $(x_i, c_i | y_i, e_i)$ in the Insurance Clause QA dataset, $x_i$ stands for the query and $c_i$ stands for the candidates with labels, while $y_i$ stands for the expected answer and $e_i$ stands for the paragraph id(s) of the ground truth. We use a standard conditional language modeling objective, and the loss function is defined as:
\begin{equation}
\begin{aligned}
    L(D_{\text{pair}}) = -\sum_i [ \log p_{LM}(y_i | x_i, c_i) \\ 
    + \log p_{LM}(e_i | x_i, c_i) ]
\end{aligned}
\end{equation}

After the Insurance Clause QA fine-tuning, the models acquired proficiency not only in giving detailed explanations of the clause text, but also in distinguishing the most relevant parts of the retrieved results and excluding incorrect information. This dual-objective training approach enhances the model's overall performance and robustness in handling complex Insurance Clause queries.

\subsection{Implementation Details}

We show our implementation details in the model fine-tuning stage here.
All experiments are conducted using 3 L40s(48GB). We use Llama-Factory as the fine-tune framework. The hyperparameters for the LoRA fine-tuning is shown in Table \ref{lora cs}, \ref{lora database} and \ref{lora clause}.
\begin{table}[h!]
    \centering
    \caption{Hyperparameters for the LoRA setting: Commonsense QA task.}

    \begin{tabular}{cc}
        \toprule
        \textbf{Task} & \textbf{Commonsense QA} \\
        \midrule
        train batch size & 4 \\
        train epoch & 5 \\
        cutoff length & 1024 \\
        LoRA alpha & 16 \\
        r & 8 \\
    \bottomrule
    \end{tabular}
    \label{lora cs}
\end{table}

\begin{table}[h!]
\centering
    \caption{Hyperparameters for the LoRA setting: Database QA task.}

    \begin{tabular}{cc}
        \toprule
        \textbf{Task} & \textbf{Database QA} \\
        \midrule
        train batch size & 1 \\
        train epoch & 3 \\
        cutoff length & 3072 \\
        LoRA alpha & 16 \\
        r & 8 \\
        \bottomrule
    \end{tabular}
    \label{lora database}
\end{table}

\begin{table}[h!]
\centering
    \caption{Hyperparameters for the LoRA setting: Clause QA task.}
    \begin{tabular}{cc}
        \toprule
        \textbf{Task} & \textbf{Clause QA} \\
        \midrule
        train batch size & 4 \\
        train epoch & 3 \\
        cutoff length & 2048 \\
        LoRA alpha & 16 \\
        r & 8 \\
        \bottomrule
    \end{tabular}
    \label{lora clause}
\end{table}

\clearpage

\section{Demo and Code}
For Qwen1.5-14B-Chat, one of the models used in our experiments, we created online demos using the fine-tuned version of the model, which we refer to as \emph{InsLLM}. These demos showcase the performance of the model fine-tuned on the InsQABench dataset. The online demos, along with the source code, are available at the following URL: \href{https://github.com/HaileyFamo/InsQABench.git}{InsQABench}.

\begin{figure*}[htbp]
    \centering
    \includegraphics[width=\textwidth]{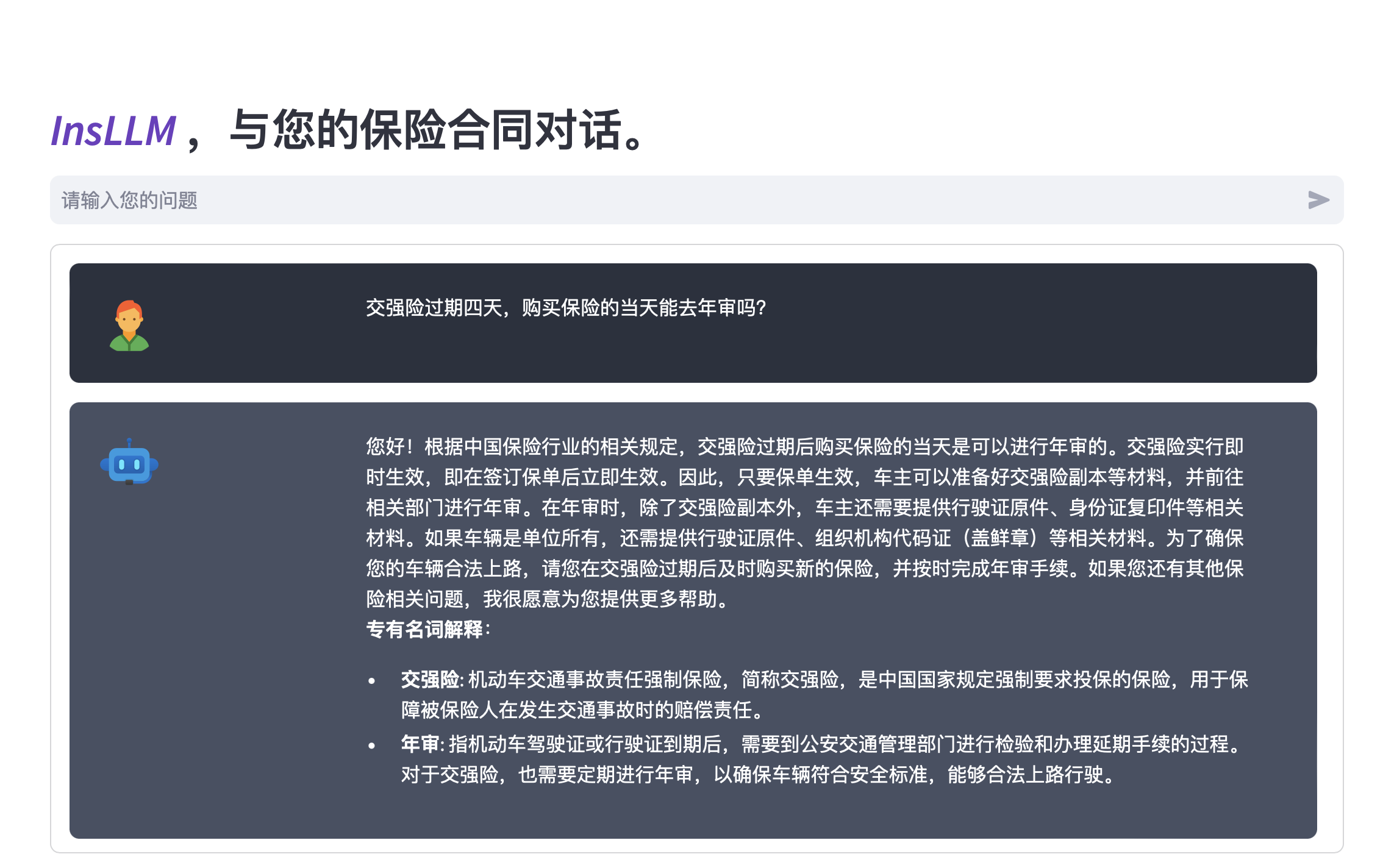}
    \caption{A demo of InsLLM for the Commonsense QA task.}
\end{figure*}

\vspace{10pt}

\begin{figure*}[htbp]
    \includegraphics[width=\textwidth]{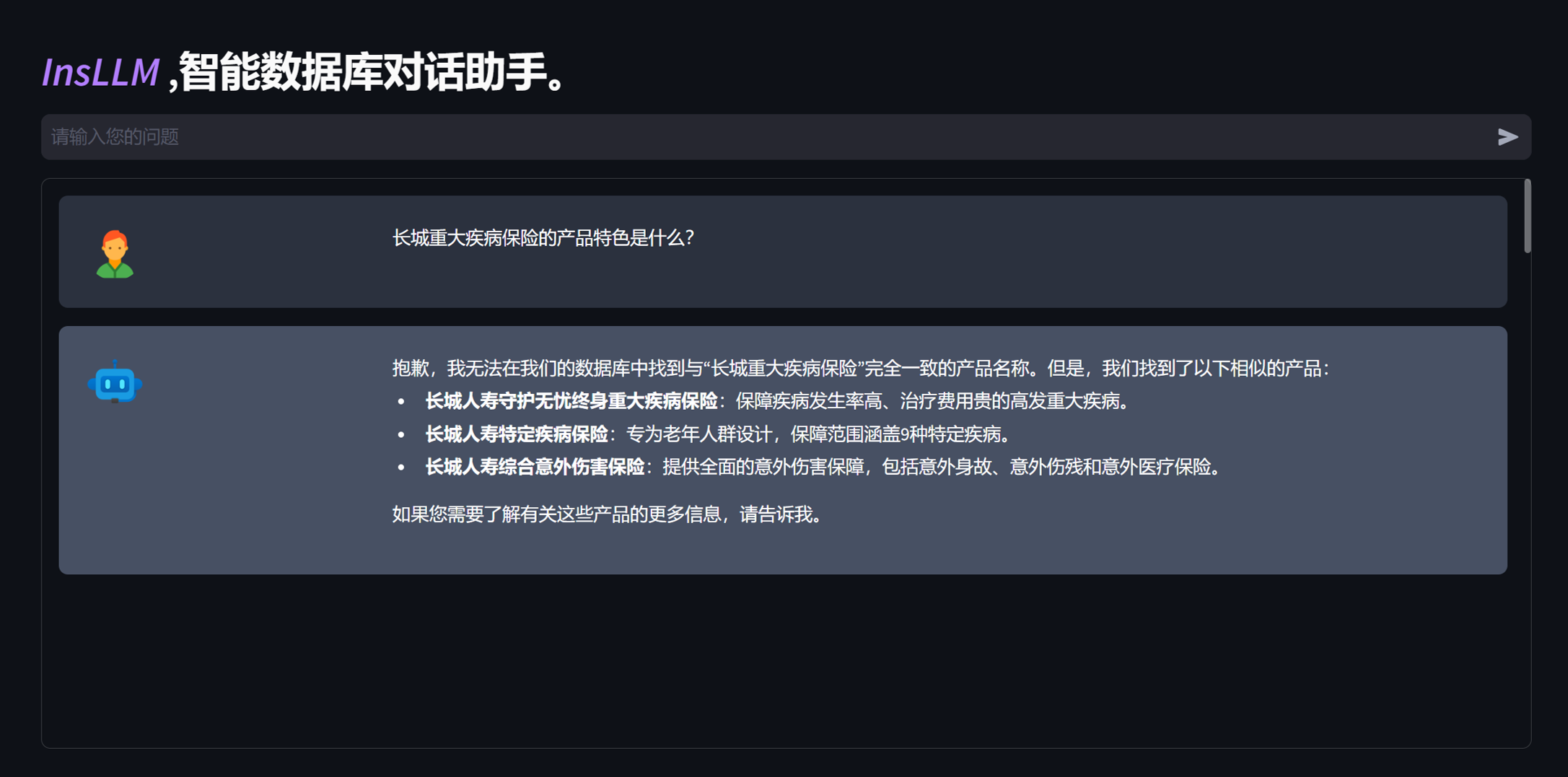}
    \caption{A demo of InsLLM for the DB QA task.}
\end{figure*}

\vspace{10pt}

\begin{figure*}[htbp]
    \includegraphics[width=\textwidth]{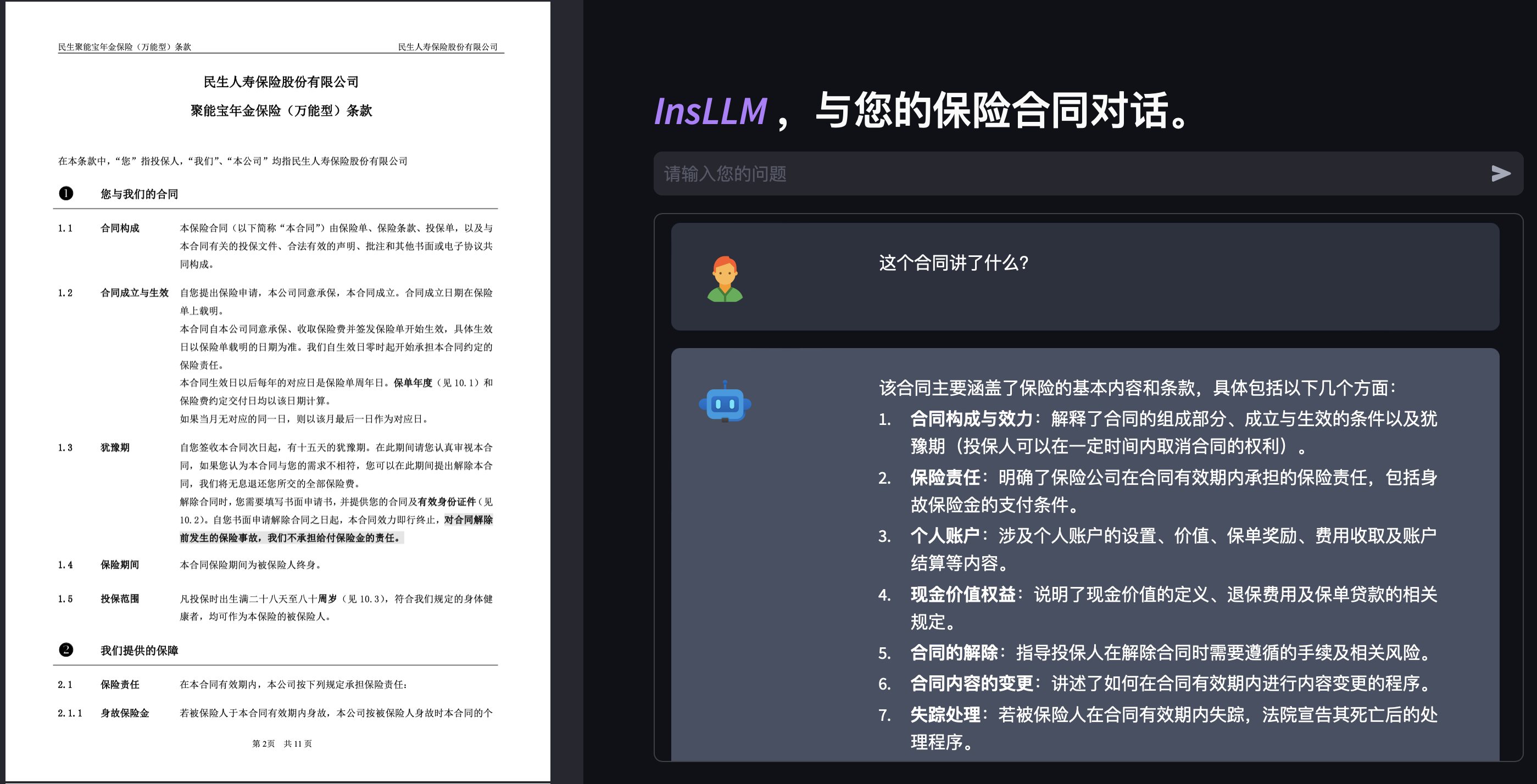}
    \caption{A demo of InsLLM for the Clause QA task.}
\end{figure*}

\clearpage
\section{Prompt Engineering}
We present the prompts used here.

\begin{figure*}[htbp]
    \centering
    \begin{tcolorbox}[colframe=black, colback=white]

    \begin{center}
    \textbf{Generation Prompt for the SQL statement}
    \end{center} 

    \textbf{Task Definition:} \\
   You are a database engineer in the insurance field, now let you analyze the user's Question, get the Thought process Thought, and write the SQL statement to solve the problem. I'll give you information about the tables in the database and the history of your previous answers. The answer history includes the previous Thought process, the SQL statement generated by the Thought process, and the results retrieved from the database using the SQL statement.\\
   Here is information about the database tables you might use :\{table\_info\}
    
    \textbf{Requirements:} 
    \begin{itemize}
        \item You need to determine whether the previous query results can answer the user's question. If the previous query results can not get relevant information, then you can not easily answer no answer, but need to continue to perform the Thought process, try to modify the previous SQL statement. so that the SQL query can find the answer from the database as much as possible. If the previous results are sufficient to answer the Question in question, then you don't need to go through the Thought process and simply answer "Ready!" You don't have to explain why.
        \item Please do not answer Ready quickly, make sure you have done enough Thought and SQL statement generation process before answering Ready.
        \item You don't want to write the entire SQL statement in one step, you should analyze the current problem, solve it step by step, and think about how the next SQL statement should be written. This SQL statement can be a step towards solving the intermediate problem. I expect a full thought process and SQL statements.
        \item The generated SQL statement is the Mysql standard, only one SQL statement can be generated at a time, and should be written in a line, you do not need to explain.
        \item SQL statements are wrapped in <sql></sql>, and execution results are wrapped in <exe></exe>.
    \end{itemize}

    \textbf{Expected Output:} \\
    <sql>...</sql> or Ready!
    \end{tcolorbox}
    \caption{The generation prompt for the SQL statement.}
    \label{fig:clause gen prompt}
\end{figure*}

\begin{figure*}[htbp]
    \centering
    \begin{tcolorbox}[colframe=black, colback=white]

    \begin{center}
    \textbf{Generation Prompt for the Database QA Answers}
    \end{center} 

    \textbf{Task Definition:} \\
   You as an insurance industry customer service, now give you a user's question and query process from the database, please answer the user's question in a customer service tone. Thought is a step by step thinking process for a problem, <sql></sql> wraps the SQL statement to be executed according to this thinking, <exe></exe> wraps the query result obtained from executing the SQL statement in the previous row in the database. Please answer the questions based on this series of reflections and inquiry results. Do not answer questions that are not related to the Question.\\
    Here is information about the database tables you might use :\{table\_info\}
    
    \textbf{Requirements:} 
    \begin{itemize}
        \item You should be careful to write your Answer in a customer service tone.
        \item The contents of all answers must be based on the contents of the SQLResult. You cannot Answer the Question according to your own knowledge.
        \item The Answer should be accurate and precise. For example, fuzzy matching is used in the query process, so that the results found may not be exactly the same as the product name that needs to be queried, at this time, you can not simply answer that the answer found is the answer that the user wants, you should first tell the user that we did not find the exact same product name, but we found the following similar products, and then answer subsequent questions. If the fuzzy matching is used, the query results exist exactly the same as the entity name in the user's Question, you can give a direct answer without specifying similar products. Don't answer exactly according to the above content, just express the exact corresponding meaning.
    \end{itemize}

    \textbf{Expected Output:} \\
    Answer:
    \end{tcolorbox}
    \caption{The generation prompt for the Database QA Answer pair.}
    \label{fig:clause gen prompt}
\end{figure*}

\begin{figure*}[htbp]
    \centering
    \begin{tcolorbox}[colframe=black, colback=white]

    \begin{center}
    \textbf{Generation Prompt for the Insurance Clause QA Pairs}
    \end{center} 

    \textbf{Task Definition:} \\
    Simulate the following conversation scenario: A Chinese Insurance expert explains the content in an insurance contract to a consumer who is very unfamiliar with insurance. \\
    Based on the given part of the insurance contract, generate 3 questions from the consumer and corresponding answers from the expert. Each set of questions and answers is independent of each other and only has one round of conversation.\\
    
    \textbf{Requirements:} 
    \begin{itemize}
        \item The consumer's question should be based on complex reasoning from the material, but the terminology is relatively simple and does not involve more technical terms. 
        \item The expert's answer should be specific and explain in detail based on the material, in a professional but simple language that beginners can understand. At the same time, note that the expert needs to explain all the proper nouns that appear in the content.
        \item Note that all questions must be able to be answered through the material, please do not add information other than the material.
        \item The questions should be short-answer questions, please don't ask multiple-choice questions, etc.
    \end{itemize}

    \textbf{Expected Output:} \\
    List your Q and A as: \\
    \textsc{[Q1]:} ...\\
    \textsc{[A1]:} ... \\
    \textsc{[Q2]:} ...\\
    \textsc{[A2]:} ...\\
    \textsc{[Q3]:} ...\\
    \textsc{[A3]:} ...\\
    
Here is the material:
    \end{tcolorbox}
    \caption{The generation prompt for the Clause QA pairs.}
    \label{fig:clause gen prompt}
\end{figure*}

\begin{figure*}[htbp]
    \centering
    \begin{tcolorbox}[colframe=black, colback=white]
    \begin{center}
    \textbf{Evaluation Standard for the Commonsense QA Task}
    \end{center}
   
  \textbf{similarity}: \\
	The pending scored answer should be as close as possible to the reference answer in meaning. \\
	A higher similarity indicates that the answer follows the logic and core content of the reference answer closely.\\
	Answers that deviate significantly in content or meaning should receive lower scores. \\

  \textbf{accuracy}: \\
	The more similar the answer is to the reference, the more accurate it is considered. \\
	Overly general or imprecise answers should score lower in accuracy.\\
  
  \textbf{professionalism}:  \\
	The answer should demonstrate expertise in insurance, using correct terminology and providing precise explanations. \\
	However, professionalism is considered meaningful only if the answer is similar to the reference. 
    \end{tcolorbox}
    
    \begin{tcolorbox}[colframe=black, colback=white]
    \begin{center}
    \textbf{Evaluation Prompt for the Commonsense QA task}
    \end{center}
    \textbf{Task Definition:} \\
    You are a professional, impartial, and strict scorer who is qualified in Chinese Insurance field. 
	You will be given a question, a reference answer, and 4 answers from different models. \\
 
    \textbf{Requirements:}
    \begin{itemize}
        \item Please rate the pending scored answers based on the reference answer in the following aspects: \textbraceleft evaluation standard\textbraceright.
        \item Each score should be from 1(lowest)-100(highest),the minimum unit of score is 1.Your rating should be strict enough, and do not easily give full scores. In scoring, ensure differentiation among similar scores by closely comparing the detail level of answers within the same score range.
        \item In your response, you should only include a JSON object, with a python dict with keys being `model 1' to `model 4', and their values are also dict with keys being the aspects and values being the scores. Do not include any additional information or explanation. \\
    \end{itemize}

    Question: \textbraceleft question\textbraceright \\
    Reference Answer: \textbraceleft answer\textbraceright \\
    Pending scored answers from different model: \textbraceleft model answers\textbraceright
    \end{tcolorbox}

    \caption{Model-based evaluation prompt for Insurance Commensense QA task.}
    \label{fig:commonsense eval prompt}
\end{figure*}


\begin{figure*}[htbp]
    \centering
    \begin{tcolorbox}[colframe=black, colback=white]

    \begin{center}
    \textbf{Evaluation Prompt for the Database QA.}
    \end{center} 

    \textbf{Task Definition:} \\
   Now you need to act as a judge to rate the answers to the following questions in the insurance field, and I will give you a StandardAnswer to the user Question. The standard answer is correct. And then give you a response to be evaluated. Please rate this response on a scale of 0 to 100. The scoring criteria are as follows:
    
    \textbf{Requirements:} 
    \begin{itemize}
        \item Whether the answer to be evaluated is consistent with the main content of the standard answer StandardAnswer, that is, whether the key points of the answer are all there.
        \item Whether the answer to be evaluated is accurate and rigorous, that is, if the search results do not match the entity name asked by the user, whether the similarity is emphasized rather than a direct answer.
        \item Detect whether there is any behavior of making up the answer, if the answer is made up or wrong, the score should be lowered.
        \item If the answer is missed, points will be given according to the percentage of the answer to the standard answer.
    \end{itemize}

    \textbf{Expected Output:} \\
    \textsc{[<score>:"score value"]}\\
    \end{tcolorbox}
    \caption{The evaluation prompt for Insurance Database QA task.}
    \label{fig:clause gen prompt}
\end{figure*}


\begin{figure*}[htbp]
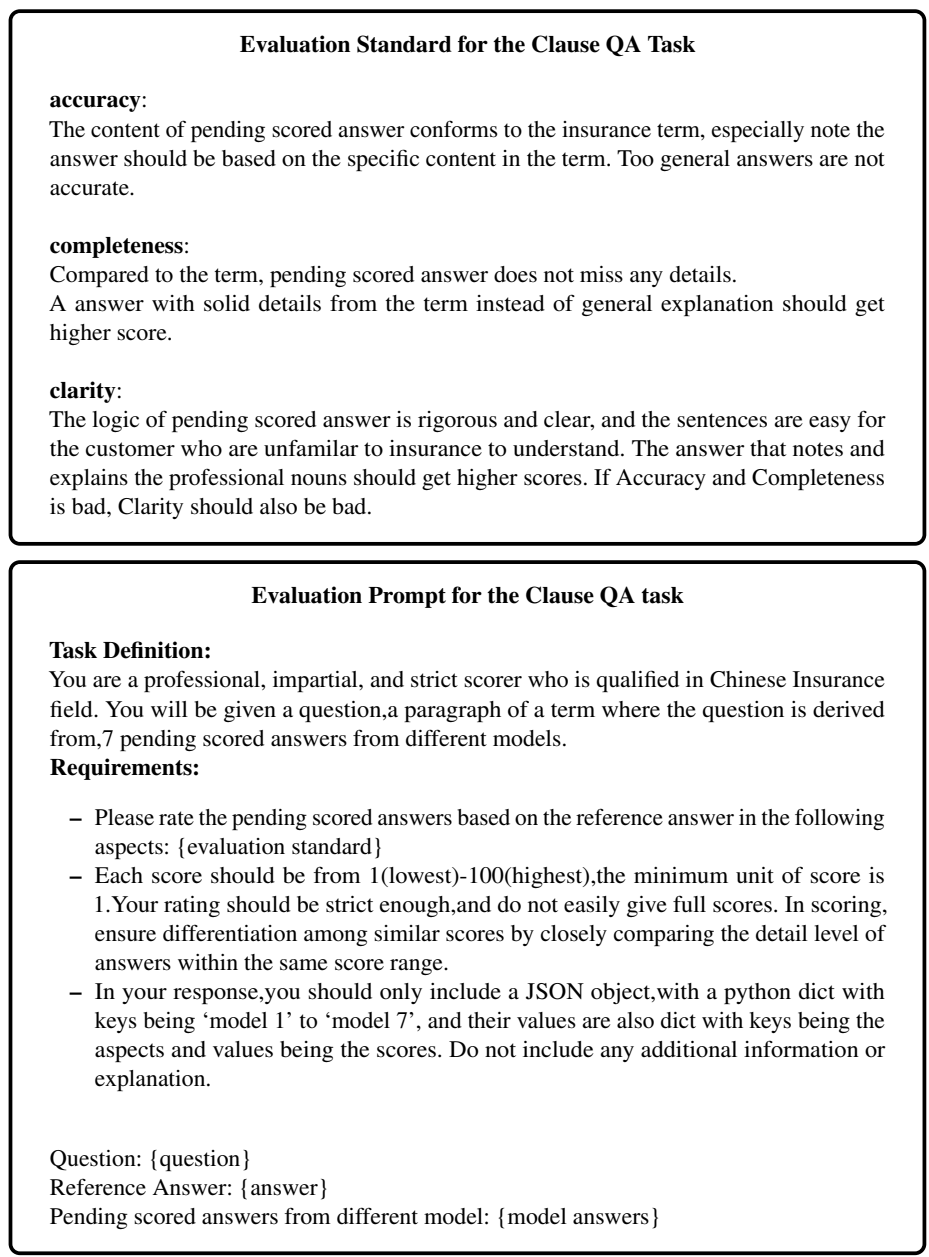

    \centering
    \begin{tcolorbox}[colframe=black, colback=white]
    \begin{center}
    \textbf{Evaluation Standard for the Clause QA Task}
    \end{center}
   
  \textbf{accuracy}: \\
  The content of pending scored answer conforms to the insurance term, especially note the answer should be based on the specific content in the term. Too general answers are not accurate.  \\

  \textbf{completeness}: \\
	Compared to the term, pending scored answer does not miss any details. \\
	A answer with solid details from the term instead of general explanation should get higher score. \\
  
  \textbf{clarity}:  \\
  The logic of pending scored answer is rigorous and clear, and the sentences are easy for the customer who are unfamilar to insurance to understand. The answer that notes and explains the professional nouns should get higher scores.
  If Accuracy and Completeness is bad, Clarity should also be bad.
    \end{tcolorbox}
    
    \begin{tcolorbox}[colframe=black, colback=white]
    \begin{center}
    \textbf{Evaluation Prompt for the Clause QA task}
    \end{center}
    \textbf{Task Definition:} \\
    You are a professional, impartial, and strict scorer who is qualified in Chinese Insurance field. You will be given a question,a paragraph of a term where the question is derived from,7 pending scored answers from different models.
 
    \textbf{Requirements:}
    \begin{itemize}
        \item  Please rate the pending scored answers based on the reference answer in the following aspects: \textbraceleft evaluation standard\textbraceright
        \item Each score should be from 1(lowest)-100(highest),the minimum unit of score is 1.Your rating should be strict enough,and do not easily give full scores. In scoring, ensure differentiation among similar scores by closely comparing the detail level of answers within the same score range. 
        \item In your response,you should only include a JSON object,with a python dict with keys being `model 1' to `model 7', and their values are also dict with keys being the aspects and values being the scores. Do not include any additional information or explanation. \\
     \end{itemize}
     
    Question: \textbraceleft question\textbraceright \\
    Reference Answer: \textbraceleft answer\textbraceright \\
    Pending scored answers from different model: \textbraceleft model answers\textbraceright
    \end{tcolorbox}

    \caption{Model-based evaluation prompt for Insurance Clause QA task.}
    \label{fig:clause eval prompt}
\end{figure*}

\end{document}